\documentclass[10pt,twocolumn,letterpaper]{article}

\usepackage{cvpr}              % To produce the CAMERA-READY version
\definecolor{cvprblue}{rgb}{0.21,0.49,0.74}
\usepackage[pagebackref,breaklinks,colorlinks,allcolors=cvprblue]{hyperref}

\usepackage{subcaption,amssymb,graphicx,makecell,booktabs,tabularx,multirow}
\usepackage{balance}
\definecolor{mygray}{rgb}{0.9, 0.9, 0.9}
\definecolor{darkred}{rgb}{0.5, 0.0, 0.0}
\definecolor{orange}{rgb}{1.0, 0.55, 0.0}
\definecolor{pink}{rgb}{0.984, 0.259, 0.616}

\def\paperID{2666} % *** Enter the Paper ID here
\def\confName{ICCV}
\def\confYear{2025}
\makeatletter
\renewcommand{\@fnsymbol}[1]{\ifcase#1\or \textsuperscript{\dag}\else\@arabic{#1}\fi}

\makeatother

\title{MATE: Motion-Augmented Temporal Consistency \\ 
for Event-based Point Tracking}

\author{
    Han Han, Wei Zhai\thanks{Corresponding author.}, Yang Cao, Bin Li, Zheng-jun Zha\\ % 列出所有作者
    MoE Key Laboratory of Brain-inspired Intelligent Perception and Cognition, \\
    University of Science and Technology of China \\ % 机构名称
    {\tt\small hanh@mail.ustc.edu.cn, \{wzhai056, forrest, binli, zhazj@ustc.edu.cn\}} % 邮箱
}

\begin{document}

\maketitle
\begin{abstract}
Tracking Any Point (TAP) plays a crucial role in motion analysis. Video-based approaches rely on iterative local matching for tracking, but they assume linear motion during the blind time between frames, which leads to point loss under large displacements or nonlinear motion. The high temporal resolution and motion blur-free characteristics of event cameras provide continuous, fine-grained motion information, capturing subtle variations with microsecond precision. This paper presents an event-based framework for tracking any point, which tackles the challenges posed by spatial sparsity and motion sensitivity in events through two tailored modules. Specifically, to resolve ambiguities caused by event sparsity, a motion-guidance module incorporates kinematic vectors into the local matching process. Additionally, a variable motion aware module is integrated to ensure temporally consistent responses that are insensitive to varying velocities, thereby enhancing matching precision.
To validate the effectiveness of the approach, two event dataset for tracking any point is constructed by simulation. The method improves the $Survival_{50}$ metric by 17.9\% over event-only tracking of any point baseline. Moreover, on standard feature tracking benchmarks, it outperforms all existing methods, even those that combine events and video frames.
% To validate the effectiveness of the approach, two event dataset for tracking any point is constructed by simulation, and is applied in experiments together with two real-world datasets. The experimental results show that the proposed method outperforms existing SOTA methods. Moreover, it achieves 150\% faster processing with competitive model parameters.
\end{abstract}
    
\section{Introduction}
\label{sec:intro}

% Long-term point tracking aims to determine the projection of the same physical point across different frames, which is crucial for understanding motion in the real world, such as autonomous driving and robotic navigation \cite{zhao2022particlesfm,chen2024leap,smith2024flowmap}. This technique generates point trajectories by temporally matching pixel appearance features from consecutive frames \cite{doersch2022tap, doersch2023tapir, harley2022particle, weikang2023context, zheng2023pointodyssey}. Consequently, it estimates motion during the blind time between frames, leading to suboptimal results under rapid and nonlinear motion. 
\begin{figure}[tp]
    \centering
    \includegraphics[width=\linewidth]{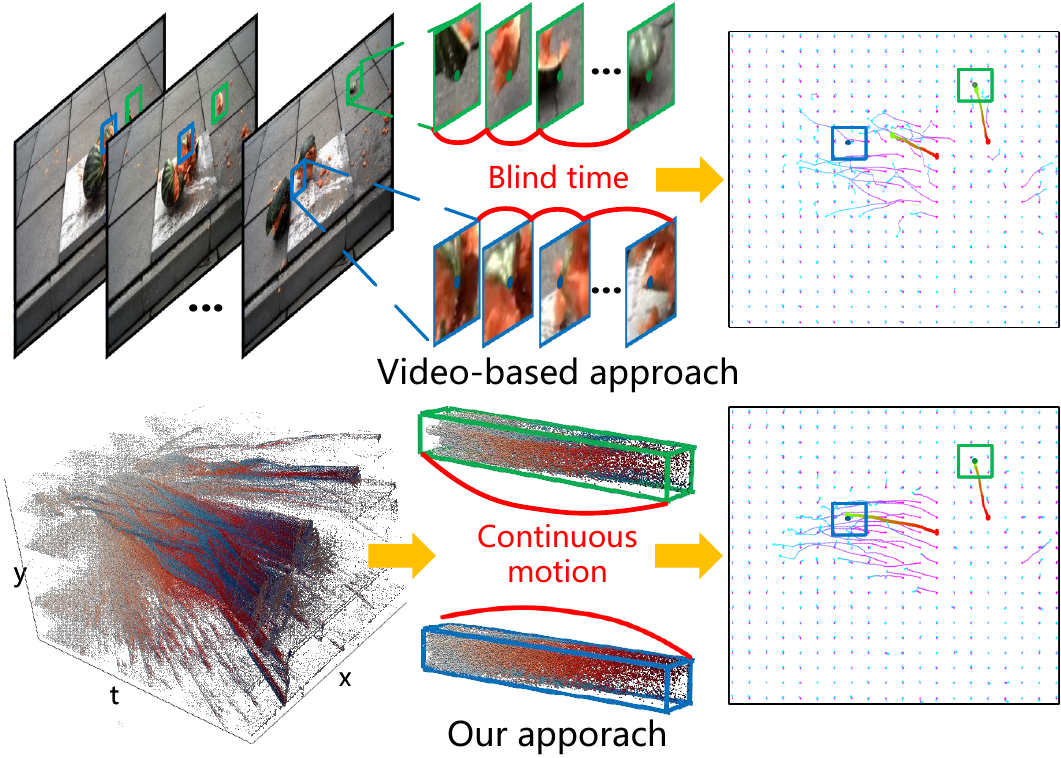}
    \caption{Video-based point tracking method (first row) face limitations in tracking objects with varying motion states, primarily due to their reliance on slow linear motion assumptions during blind times. Our approach (second row) leverages continuous motion information from events, achieving smoother and more accurate results.}
    \label{fig:rgbvsevent}
\end{figure}
Tracking Any Point (TAP) aims to determine the subsequent positions of a given query point on a physical surface over time, which is essential for understanding object motion in the scene. It becomes even more vital for autonomous driving and embodied agents \cite{zhao2022particlesfm,chen2024leap,smith2024flowmap}, where operations require precise spatial control of objects over time.
\begin{figure*}[!t]
    \centering
    \begin{subfigure}{0.46\linewidth}
        \includegraphics[width=\linewidth]{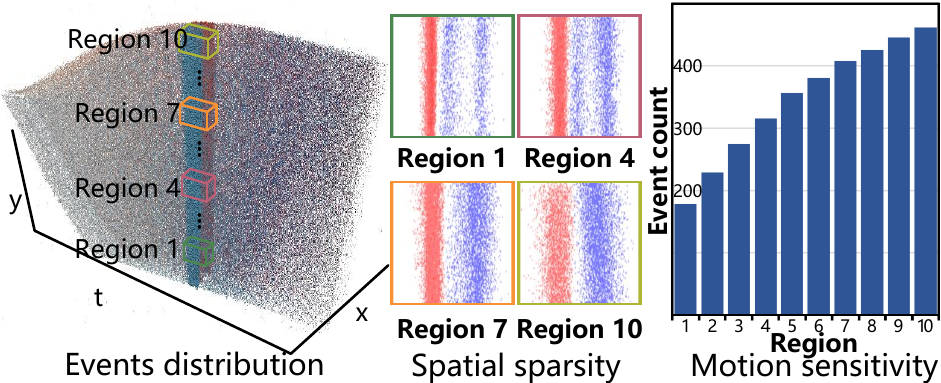}
        \caption{The spatio-temporal properties of events.}
        \label{fig:stick}
    \end{subfigure}
    \hspace{0.05\linewidth} 
    \begin{subfigure}{0.46\linewidth}
        \includegraphics[width=\linewidth]{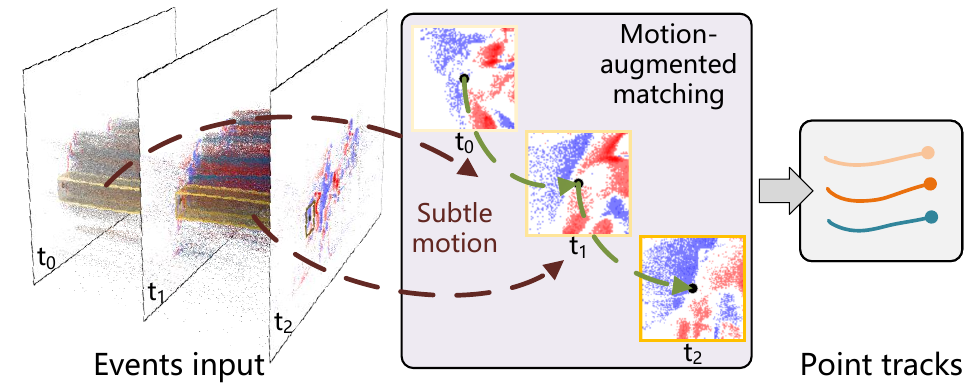}
        \caption{The solution of this paper.}
        \label{fig:solution}
    \end{subfigure}    
    \caption{(a) The spatio-temporal distribution of events generated by a stick rotating uniformly around a pivot. Sampling from different regions along the stick at fixed time intervals reveals spatial sparsity. The event count increases with distance from the pivot, corresponding to higher linear velocity. This indicates that faster motion triggers more events, and vice versa. (b) This paper leverages the temporal continuity of events to mitigate spatial sparsity and exploits subtle motion to guide matching, ultimately improving tracking accuracy.}
\end{figure*}

Recent methods rely on video input, predicting the positions of query points by matching their appearance features with local regions in subsequent frames \cite{doersch2022tap, doersch2023tapir, harley2022particle}.
However, as these methods assume slow linear motion during the blind time between frames, they face the challenge of objects may undergo large displacements or nonlinear motion, causing query points to exceed the bounds of local regions and resulting in ambiguities in feature matching, see \cref{fig:rgbvsevent}. While some approaches attempt to mitigate this by considering spatial context \cite{weikang2023context, cho2024local}, they still struggle to overcome the lack of motion during blind time.

To cope with the above issue, this paper utilize event cameras to capture the motion during blind time.
Event cameras are bio-inspired sensors that respond to pixel-level brightness changes with microsecond temporal resolution, generating sparse and asynchronous event streams \cite{lichtsteiner2008128,wu2024event}. They have the characteristics of high temporal resolution, no motion blur, and low energy consumption. Therefore, parsing motion during blind times using event streams is a feasible solution for tracking any point. Furthermore, capturing motion alone significantly reduces computational overhead compared to traditional frame-based cameras, enabling more efficient methods.

Recent methods \cite{liu2024tracking, hamann2024event} have attempted to use event cameras for tracking any point. However, they still rely solely on appearance-based matching for tracking. As shown in \cref{fig:stick}, the unique spatio-temporal properties of events make existing methods difficult to apply, manifesting in two ways: 1) Event cameras respond only to pixels with brightness changes, resulting in a sparse spatial distribution that introduces ambiguities in appearance-based matching due to spatial discontinuities.
2) The variable speed of scene objects causes fluctuations in event update frequencies, impacting the temporal consistency of event representations. Higher motion speeds lead to increased event update frequencies and densities, and vice versa. Density variations cause inconsistencies in event representations over time, affecting the precision and reliability of temporal matching.

To address these issues, this study leverages the temporal continuity of events to guide matching, alleviating ambiguities caused by spatial sparsity, as illustrated in \cref{fig:solution}. The temporal dynamics of events capture subtle variations in motion trajectory, complementing spatial appearance to provide coherent matching cues. Additionally, by modeling kinematic features to estimate object speed and motion patterns, the method dynamically adjusts appearance feature extraction, ensuring temporally consistent responses.

% since events model scene structure but struggle with texture, applying appearance features solely for temporal matching causes ambiguities.To eliminate this, 
% Besides,  
Therefore, this paper proposes a novel framework that utilizes \textbf{M}otion-\textbf{A}ugmented \textbf{T}emporal consistency for \textbf{E}vent-based point tracking (\textbf{MATE}). It includes a Motion-Guidance Module (MGM) and a Variable Motion Aware (VMA) module.
Specifically, MGM leverages the gradient from event time surface to compute kinematic features that construct a dynamic-appearance space, clarifying ambiguities in feature matching and guiding the extraction of temporally consistent appearance features. Additionally, to model the non-stationary states of events, VMA is introduced to employ kinematic cues for adaptive correction and generate temporally consistent responses by combining long-term memory parameters with hierarchical appearance.
The method is evaluated across four datasets on two tasks, with experimental results demonstrating its superiority.
The main contributions of this paper can be summarized as follows:
\begin{itemize}
    % \item An event-based framework for tracking any point is presented to monitor points on the surfaces of high-speed moving objects.
    \item An event-based framework for tracking any point, named MATE, is presented to monitor surface points on objects by leveraging the temporal continuity of events.
    % \item This work reveals the impact of spatial sparsity and motion sensitivity in event data on TAP. To address these limitations, a motion guided module is proposed to mitigate sparsity-induced ambiguities, while a variable motion aware module models non-stationary event state, enhancing tracking accuracy. 
    \item This paper reveals the impact of spatial sparsity and motion sensitivity in events on TAP. To address these limitations, a motion-guidance module is proposed to enhance matching process using temporal continuity, while a variable motion aware module models kinematic cues to estimate non-stationary event states, improving performance.
    \item Experimental results demonstrate that the proposed approach outperforms current state-of-the-art methods in both tracking of any point and feature tracking tasks across four datasets, with fewer model parameters.
\end{itemize}

% \begin{figure}[!t]
%     \centering
%     \begin{subfigure}{\linewidth}
%         \includegraphics[width=\linewidth]{figs/stick.pdf}
%         \caption{The spatio-temporal properties of events.}
%         \label{fig:stick}
%     \end{subfigure}
%     \begin{subfigure}{\linewidth}
%         \includegraphics[width=\linewidth]{figs/stick 3.pdf}
%         \caption{The solution of this paper.}
%         \label{fig:motivation}
%     \end{subfigure}    
%     \caption{(a) The spatio-temporal distribution of events generated by a stick rotating uniformly around a pivot. Sampling from different patches along the stick reveals spatial sparsity of events. Counting the events at each patch shows a positive correlation between event update frequency and object speed, which causes temporal inconsistencies with varying motion speeds. (b) This paper leverages the temporal continuity of events to capture subtle motion changes, enhancing the matching process and thereby improving tracking accuracy.}
% \end{figure}

\section{Related Work}
\label{sec:related}

\subsection{Event-based Optical Flow}
Event-based optical flow estimation can be categorized into model-based and learning-based methods.
Model-based methods rely on physical priors \cite{nagata2023tangentially, almatrafi2020distance, hamann2024motion, shiba2024secrets}, primarily using contrast maximization to estimate optical flow by minimizing edge misalignment. Unfortunately, the strict motion assumptions inherent in these methods struggle in complex scenes, leading to decreased accuracy.
% Optical flow captures motion within a time interval in the field of view. The high dynamic range and low latency of event cameras make them a viable solution for optical flow.   

Learning-based methods have significantly improved the quality of optical flow estimation \cite{zhu2018ev, gehrig2024dense, wan2024event, zhuang2024ev, zhu2019unsupervised, wan2022learning, wan2025emotiveeventguidedtrajectorymodeling}.
They can be classified into unsupervised \cite{zhu2018ev, zhu2019unsupervised, zhuang2024ev} and supervised approaches \cite{ding2022spatio, gehrig2024dense, wan2024event, wan2022learning}. For the former, several methods have been proposed, including those that use event cameras alone \cite{zhu2019unsupervised, zhuang2024ev} or in combination with other data modalities \cite{zhu2018ev}. They employ motion compensation to warp and align data across time to construct a loss function. The latter commonly utilize a coarse-to-fine pyramid structure or iterative optimization to refine the estimation. For example, Gehrig \etal \cite{gehrig2021raft} propose E-RAFT, which mimics iterative optimization algorithms by updating the correlation volume to optimize optical flow. 
% Zhu \etal \cite{zhu2018ev} proposed the first unsupervised optical flow estimation method that utilized geometric constraints among optical flow, depth, and camera pose to construct a loss function. However, this method relies on the presence of grayscale images, which struggle in blurred scenes. To address this, several works \cite{zhu2019unsupervised, zhuang2024ev} introduced a motion-compensation loss that sharpened the edges of warped event images, allowing the network to learn without RGB images. In contrast, Gehrig \etal \cite{gehrig2019end} enhanced \cite{zhu2018ev} with a learnable event map, solving optical flow through supervised learning. To handle scenes of varying scales, Wan \etal \cite{wan2022learning} and Li \etal \cite{li2021lightweight} refined optical flow estimation using a coarse-to-fine pyramid structure. Instead of this approach, Gehrig \etal \cite{gehrig2021raft} proposed E-RAFT, which mimics iterative optimization algorithms by updating the correlation volume to optimize optical flow.

Although these methods have achieved promising results, challenges remain when applying them to TAP. Since optical flow is computed over neighboring time, linking motion vectors over long time leads to error accumulation.
% adversely affecting long-term tracking accuracy.
Additionally, optical flow calculates the correspondence between pixel points, rather than physical surface points.
\begin{figure*}[!t]
    \centering
    \includegraphics[width=\linewidth]{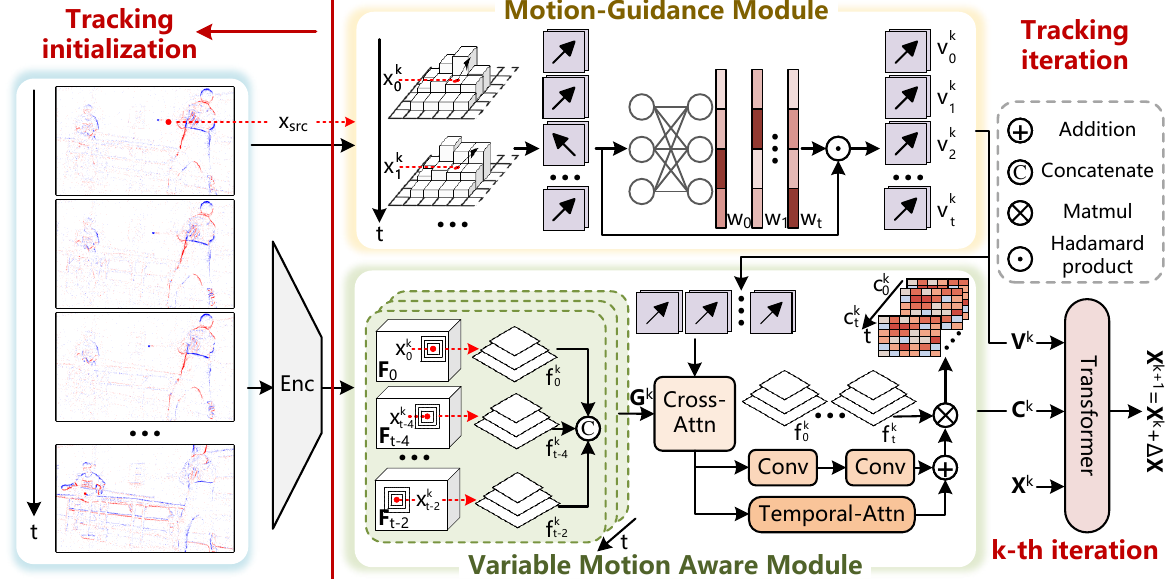}
    \caption{MATE Overview. At the initial stage, given query points and event data, the encoder extracts feature representations $\mathbf{F}$. At the $k$-th iteration, motion guidance module derives kinematic vectors $\mathbf{V}^k$ from the time surface to guide appearance matching in variable motion aware module and enrich the correlation map, which is fed into the transformer to predict the next point location. Variable motion aware module extracts hierarchical features from $\mathbf{F}$, utilizes $\mathbf{V}^k$ as guidance to compute the correlation map $\mathbf{C}^k$. The transformer takes Point position $\mathbf{X}^k$, kinematic vectors $\mathbf{V}^k$ and correlation map $\mathbf{C}^k$ as input to compute the displacement $\Delta \mathbf{X}$ for the next iteration, and this process repeats until the final trajectory is obtained.}
    \label{fig:network}
\end{figure*}

\subsection{Event-based Feature Tracking}
Feature tracking aims to predict the trajectories of keypoints.
% Due to the low latency and high dynamic range of event cameras, researchers are increasingly focused on them in feature tracking within high-speed dynamic scenes.
Early approaches can be grouped into two types: one \cite{kueng2016low, zhu2017event} treats feature points as event sets and tracks them using the ICP \cite{best1992method} method, while the other \cite{gehrig2020eklt} extracts feature blocks from reference frames and matches them by computing brightness increments from events. Moreover, event-by-event trackers \cite{alzugaray2018ace, alzugaray2020haste, hu2022ecdt, wang2024asynchronous} exploit the inherent asynchronicity of events. Unfortunately, these methods involve complex parameters that require extensive manual tuning for different event cameras and new environments.

To tackle these deficiencies, learning-based feature tracking methods have gained attention from researchers \cite{messikommer2023data, li20243d}. DeepEvT \cite{messikommer2023data} is the first data-driven method for event feature tracking. \cite{li20243d} expands 2D feature tracking to 3D and collected the first event 3D feature tracking dataset.

However, existing methods track high-contrast points by relying on local feature descriptors. TAP requires the ability to track points in low-texture areas, where current methods struggle to establish reliable descriptors.

\subsection{Tracking Any Point}
Video-based tracking of any point methods \cite{doersch2022tap, doersch2023tapir, harley2022particle, cho2024local, zheng2023pointodyssey} have been developed for a few years.
These methods model the appearance around the points, using MLPs to capture long-range temporal contextual relationships across frames. During inference, they employ a sliding time window to handle long videos.
For example, PIPs \cite{harley2022particle} frames pixel tracking as a long-range motion estimation problem, updating trajectories through iterative local searches. In contrast, TAP-Net \cite{doersch2022tap} formalizes the problem as tracking any point, overcoming occlusion through global search. However, these methods track points independently, leading to ambiguities in feature matching. Consequently, some studies \cite{weikang2023context, cho2024local, karaev23cotracker, zhai2024background, zhai2023exploring} have been proposed to utilize spatial context to alleviate this issue.
In addition to methodological innovations, the PointOdyssey \cite{zheng2023pointodyssey} and the TAP-Vid-Davis \cite{doersch2022tap} datasets are collected to advance the field and are used as training data or benchmark datasets for evaluation.

Event-based tracking of any point methods \cite{liu2024tracking, hamann2024event} have only been proposed in the past year. FE-Tap \cite{liu2024tracking} performs tracking of any point by fusing video frames and events, while ETAP \cite{hamann2024event} constructs a backward event flow and optimizes network parameters using contrastive loss. 

Although these methods have achieved promising results, they all rely on appearance features for correspondence matching. However, the spatial sparsity of events leads to misalignment when solely modeling based on appearance. Additionally, the temporal inconsistency in event density, caused by variable-speed motion, affects temporal context modeling. In this paper, a motion-guidance module is designed to construct a dynamic-appearance matching space, thereby reducing matching ambiguity. Moreover, a variable motion aware module is employed to generate temporally consistent responses for correlation operations. The method is trained and tested on simulated event modalities derived from the \cite{zheng2023pointodyssey} dataset.

\section{Method}
\label{sec:method}
\subsection{Setup and Overview}
Tracking any point process typically involves two stages: initializing tracking points and features, and iteratively updating point positions and associated features. Following this pipeline, an event-based method is proposed for tracking any point, as shown in \cref{fig:network}. 

% let $E_j=\left\{\left(x_k, y_k, t_k, p_k\right)\right\}_{k=1}^N $ denote the event stream from $t_{j-1}$ to $t_{j}$, where $N$ is the number of events, and each event is a 4-tuple consisting of the coordinates $x_k$ and $y_k$, timestamp $t_k$, and polarity $p_k \in \{-1, +1\}$. 
Given a query point $x_{src}\in \mathbb{R}^2$ at the initial time, subsequent events are represented by the Time Surface (TS) \cite{mueggler2017fast}, where each pixel records the timestamp of the most recent event, capturing motion process over a period of time. During the initialization period, the TS representation is fed into a residual network \cite{he2016deep} to extract features $\mathbf{F}^{T \times C \times H \times W} = \left\{\mathbf{F}_0,\mathbf{F}_1,...,\mathbf{F}_T\right\}$, where $T,C,H,W$ represent the number of time steps, channel size, height, and width of feature maps, respectively. 

At the iterative stage, let $x_t^k$ denote the coordinate of the query point at time $t$ after the $k$-th iteration. The MGM estimaties the parameters of the local timestamp plane around $x_t^k$, yielding the kinetic vector $v_t^k$, which encodes the local motion tendency. To compute the displacement of coordinate, the VMA module samples local feature patches from $\mathbf{F}$ at the coordinates $\{x_0^k, x_{t-4}^k, x_{t-2}^k\}$, constructing feature pyramids $\{f_0^k, f_{t-4}^k, f_{t-2}^k\}$. These patches are aligned under the guidance of $v_t^k$ and used to compute the correlation map $c_t^k$ with the feature at $x_t^k$. The complete correlation set over all time steps is denoted as $\mathbf{C}^k = \{c_0^k,c_1^k,...,c_t^k\}$. The kinetic vectors $\mathbf{V}^k$, correlation maps $\mathbf{C}^k$, and point positions $\mathbf{X}^k$ are processed by a transformer to predict the displacement $ \Delta \mathbf{X} $. The updated coordinates for the next iteration are given by \cref{eq:coord_update}:
\begin{align}
\centering
    \mathbf{X}^{k+1} = \mathbf{X}^k + \Delta \mathbf{X}.
\label{eq:coord_update}
\end{align}

% Building upon this framework, the next sections delve into the detailed workings of the MGM and VMA modules, illustrating how they contribute to the point tracking task.

\subsection{Motion-Guidance Module}
To reduce ambiguities arising from appearance-only matching, a motion-guidance module is designed to leverage the gradient characteristics of the TS to compute kinematic vectors, thereby building a dynamic-appearance matching space and subsequently guiding the extraction of temporally consistent appearance features.

At the first iteration, the query points at all time steps are initialized as $x_{src}$. The MGM samples the TS at $x_{src}$ across all time steps to compute the the initial kinetic vectors $\mathbf{V}^0 \in \mathbb{R}^{T \times 2}$, which represent the motion trend in the $x$- and $y$-directions at each time step. At the $k$-th iteration, MGM samples the TS at $\mathbf{X}^k$ across time to update the kinetic vectors $\mathbf{V}^k \in \mathbb{R}^{T \times 2}$.
Specifically, the event stream after TS encoding is visualized as a surface in the $xyt$ spacetime domain, representing the active events surface $\Sigma_e$ \cite{benosman2013event}. The spatial gradients of this surface describe the temporal changes relative to spatial variations, establishing a derivative relationship with pixel displacement at corresponding positions, see \cref{eq:mgm_sae}
\begin{align}
\centering
    \frac{\partial \Sigma_e}{\partial x} = \left(\frac{\partial x}{\partial \Sigma_e}\right)^{-1} = \left(\frac{\partial x}{\partial t}\right)^{-1} = \frac{1}{v} .
\label{eq:mgm_sae}
\end{align}
For the query point $x_t^k$, this paper treats the surface formed by neighboring pixels as a surface of active events.
Using the coordinates and of neighboring points $(x_1,y_1,t_1)$, $(x_2,y_2,t_2)$, ..., a system in \cref{eq:equations} is constructed: 
\begin{align}
\centering
\begin{bmatrix}
    x_1&y_1&t_1&1\\
    x_2&y_2&t_2&1\\
    \vdots&\vdots&\vdots&\vdots
\end{bmatrix}
\begin{bmatrix}
    a\\
    b\\
    c\\
    d
\end{bmatrix}
= 0 .
\label{eq:equations}
\end{align}
Here, $(a,b,c,d)$ are coefficients for the tangent plane. The spatial gradient of the surface is estimated through plane fitting using SVD, providing motion trend at the query point.
% It constructs an overdetermined systems using the $xyt$ coordinates of multiple neighboring pixels to solve for the spatial gradient of the surface, thereby obtaining the kinematic vectors at the target pixel.
However, in the presence of overlapping moving objects, the pixels at boundary intersections can exhibit multiple distinct motion states, disrupting local smoothness and resulting in deviations in the kinematic vectors. To address this, the proposed motion-guidance module employs a multi-layer perceptron to capture the temporal motion relationships, dynamically assigning weights to the kinematic vectors at different time steps, thereby correcting the motion ambiguity at the boundaries.

The corrected kinematic vectors serve two key roles: first, as input to VMA, guiding the generation of speed-insensitive appearance features for correlation; second, as inputs to the transformer, combining with correlation maps and point positions to construct a dynamic-appearance space for iterative point displacement updates.

\subsection{Variable Motion Aware Module}
Objects in the scene exhibit variable speeds in three main ways: 1) different objects move at distinct speeds, 2) the speed of a single object changes over time, and 3) different parts of the same object move at different speeds. The variation in speed affects the frequency of event updates. Assuming an object moves at speed $\mathbf{v}=(v_x, v_y)$, the illuminance change rate at any point on the edge is given by:
\begin{align}
\centering
    % \frac{dI(x,y,t)}{dt} &= \frac{\partial I}{\partial x}\frac{dx}{dt} + \frac{\partial I}{\partial y}\frac{dy}{dt} + \frac{\partial I}{\partial t} \\
    % &= \frac{\partial I}{\partial x}v_x + \frac{\partial I}{\partial y}v_y + \frac{\partial I}{\partial t},
    \frac{dI(x,y,t)}{dt} = \frac{\partial I}{\partial x}v_x + \frac{\partial I}{\partial y}v_y + \frac{\partial I}{\partial t},
\label{eq:vmam_ill}
\end{align}
where $I(x,y,t)$ represents the illuminance at position $(x, y)$ at time $t$. To simplify calculations, assuming consistent global illumination, so $\frac{\partial I}{\partial t}$ is $0$, and \cref{eq:vmam_ill} simplifies to:
\begin{align}
\centering
    \frac{dI(x,y,t)}{dt} = \frac{\partial I}{\partial x}v_x + \frac{\partial I}{\partial y}v_y = \boldsymbol{v} \cdot \nabla I .
\label{eq:vmam_ill_simplified}
\end{align}
Here, $\nabla I$ represents the spatial gradient of illuminance at the specified position $(x, y)$. 

The event generation process can be formulated as
\begin{align}
    logI(x,y,t)-logI(x,y,t-\Delta t)=pC,
\label{eq:vmam_event}
\end{align}
where $\Delta t$ is the time interval between consecutive events and $C$ is the contrast threshold of the event camera. This equation indicates that an event triggers when the logarithmic illuminance change at a location exceeds the threshold $C$. Combining \cref{eq:vmam_ill_simplified,eq:vmam_event}, it can be observed that 
\begin{align}
    f \propto \frac{\boldsymbol{v}\cdot\nabla I}{C},
\label{eq:vmam_f_v}
\end{align}
where $f$ signifies the event update frequency. Since $\nabla I$ depends only on the material properties of the object, $f$ is directly proportional to $\boldsymbol{v}$. In other words, higher speeds lead to higher event update frequencies, and vice versa.
% As shown in \cref{fig:stick}, a stick rotates around one end as the pivot. The farther a point is from the pivot, the greater its linear velocity and the more events are triggered per unit time.

Point position updates rely on consistent appearance feature matching over time. However, velocity variations disrupt the stability of event temporal distribution, causing significant matching errors. To tackle the problem, VMA is introduced to guide appearance feature matching using kinematic vectors extracted through MGM. 

Specifically, to obtain the correlation map $c_t^k$ at the $k$-th iteration and time $t$, VMA samples patch features from $\mathbf{F}_0$, $\mathbf{F}_{t-4}$, $\mathbf{F}_{t-2}$ at \{$x_0^k$, $x_{t-4}^k$, $x_{t-2}^k$\} to construct feature pyramids. After interpolation, the feature pyramid maintains consistent spatial dimensions. The features are then flattened into a one-dimensional vector and concatenated along the channel dimension to form $g_t^k \in \mathbb{R}^{1 \times C}$. Repeating this across all time steps yields $\mathbf{G}^k = \{g_0^k, g_1^k, ..., g_t^k\} \in \mathbb{R}^{T \times C}$. These features are fused with temporally contextual kinematic vectors $\mathbf{V}^k$ via cross-attention, then processed through two complementary branches: one captures short-term dependencies via two 1D temporal convolutions, while the other extracts long-term dependencies through temporal attention. The short-range and long-range features are summed together and correlated with the spatial context features of $\mathbf{X}^k$, producing $\mathbf{C}^k$. This means that the correlation map $c_t^k$ is determined by computing the correlation between the features from time steps \{$0$, $t-4$, $t-2$\} and $t$. The reason for sampling at these time is that if tracking was successful on one of these offset frames, the feature will reflect the updated appearance of the query point, yielding a more useful correlation map than the one from $\mathbf{F}_0$.

\section{Experiment}
\label{sec:exp}

\begin{figure}[t]
    \centering
    \includegraphics[width=\linewidth]{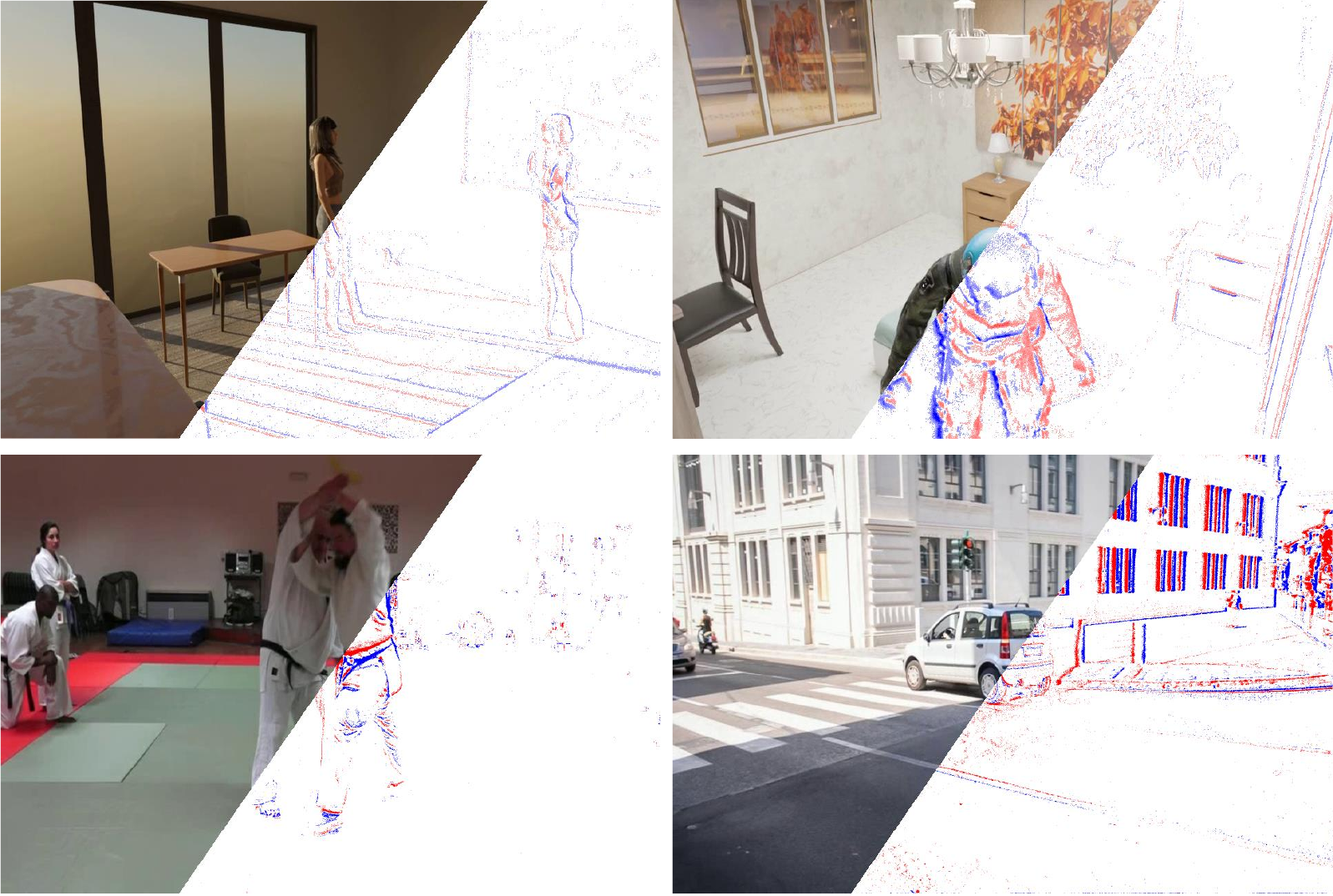}
    % \caption{Some examples from the Ev-PointOdyssey dataset. The RGB is displayed in the top-left corner, while the event modality is shown in the bottom-right corner.}
    \caption{Some examples from the Ev-PointOdyssey (top row) and Ev-Davis (bottom row) dataset. Each example displays the RGB image in the top-left corner, with the event modality visualization in the bottom-right corner.}
    \label{fig:Ev-PointOdyssey}
\end{figure}
\begin{table*}[ht]
    \centering
    \small
    \caption{The performance of the evaluated trackers on the Ev-PointOdyssey and Ev-davis datasets are reported in terms of $\delta_{avg}$, MTE and $Survival$. $\delta_{avg}$ reflects the proportion of the error between the predicted and the ground truth within a certain range. MTE represents the error between the predicted trajectory and the ground truth. $Survival$ indicates the duration of tracking. ``Ev-'' indicates that video input is replaced with event data, and the model is retrained on Ev-PointOdyssey. Best results are in bold; second-best are underlined.}
    \begin{tabular}{p{0.08\linewidth}|>{\centering\arraybackslash}p{0.068\linewidth}|>{\centering\arraybackslash}p{0.075\linewidth}>{\centering\arraybackslash}p{0.075\linewidth}>{\centering\arraybackslash}p{0.105\linewidth}|>{\centering\arraybackslash}p{0.075\linewidth}>{\centering\arraybackslash}p{0.075\linewidth}>{\centering\arraybackslash}p{0.105\linewidth}|>{\centering\arraybackslash}p{0.055\linewidth}}
    \hline
    \Xhline{2.\arrayrulewidth}
    \multirow{2}{*}{Method}        & \multirow{2}{*}{Modality} & \multicolumn{3}{c|}{Ev-PointOdyssey}         & \multicolumn{3}{c|}{Ev-Davis}  &\multirow{2}{*}{Params}                    \\  
                                    &       & $\delta_{avg}\uparrow$ & MTE $\downarrow$   & $Survival_{50} \uparrow$  & $\delta_{avg}\uparrow$ & MTE $\downarrow$   & $Survival_{16} \uparrow$  &   \\ 
    \hline
    \Xhline{2.\arrayrulewidth}

    \multicolumn{1}{l|}{PIPs++ \cite{zheng2023pointodyssey}}       & \multirow{3}{*}{Video}  & 0.336             & 26.95  & 0.505        & 0.697    & 4.16   & 0.897      & 17.6M            \\
    \multicolumn{1}{l|}{Cotracker \cite{karaev23cotracker}} &                    & 0.302             & - & \underline{0.552} & 0.726    & 3.99   & 0.902      & 24.2M            \\ 
    \multicolumn{1}{l|}{Cotracker3 \cite{karaev2024cotracker3}} &       & \textbf{0.364} & \underline{26.84}& 0.409 & \underline{0.739}    & \underline{3.73}   & \underline{0.913}      & 25.4M            \\ \hline 
    
    \multicolumn{1}{l|}{Ev-PIPs++ \cite{zheng2023pointodyssey}} & \multirow{2}{*}{Event} & 0.306             & 57.36 & 0.475 & 0.668    & 5.21   & 0.874      & 24.2M            \\ 
    \multicolumn{1}{l|}{Ev-Cotracker3 \cite{karaev23cotracker}} &                        & 0.316             & 42.55 & 0.428 & 0.696    & 4.52   & 0.887      & 25.4M            \\ \hline

    \rowcolor{mygray}
    \multicolumn{1}{l|}{Ours}         & Event                 & \underline{0.362} & \textbf{25.63} & \textbf{0.560}  & \textbf{0.747}    & \textbf{3.47}   & \textbf{0.928}      & \textbf{6.6M}     \\ \hline \Xhline{2.\arrayrulewidth}
    \end{tabular}    
    \label{tab:ev_pointodyssey}
\end{table*}
\begin{table}[t]
    \centering
    \small
    \caption{The performance of different feature tracking methods on the EC and EDS datasets. The proposed approach achieves the best performance on both datasets, with particularly notable improvements on the EDS dataset, which involves more camera motion. ``E+V'' represents events and videos.}
    \resizebox{\columnwidth}{!}{
        \begin{tabular}{p{0.11\linewidth}|>{\centering\arraybackslash}p{0.08\linewidth}|>{\centering\arraybackslash}p{0.075\linewidth}>{\centering\arraybackslash}p{0.1\linewidth}|>{\centering\arraybackslash}p{0.075\linewidth}>{\centering\arraybackslash}p{0.1\linewidth}}
        \hline
        \Xhline{2.\arrayrulewidth}
        \multirow{2}{*}{Method}   & \multirow{2}{*}{Input}     & \multicolumn{2}{c|}{EDS}    & \multicolumn{2}{c}{EC}   \\  
                                  &      & FA $\uparrow$       & EFA $\uparrow$   & FA $\uparrow$      & EFA $\uparrow$  \\ 
        \hline
        \Xhline{2.\arrayrulewidth}
        \multicolumn{1}{l|}{PIPs++ \cite{zheng2023pointodyssey}}         & \multirow{2}{*}{Video} & 0.675     & 0.580          & 0.696    & 0.675         \\ 
        \multicolumn{1}{l|}{Cotracker3 \cite{karaev2024cotracker3}}      & & 0.630     & 0.572          & 0.648    & 0.641         \\ \hline
        \multicolumn{1}{l|}{EKLT   \cite{gehrig2020eklt}}           & \multirow{3}{*}{E+V} & 0.325     & 0.205          & 0.811    & 0.775         \\
        \multicolumn{1}{l|}{DeepEvT\cite{messikommer2025data}}      &           & 0.576     & 0.472          & 0.825    & 0.818         \\ 
        \multicolumn{1}{l|}{FE-TAP \cite{liu2024tracking}}          &           & 0.676     & 0.589          & 0.844    & 0.838         \\ \hline
        
        \multicolumn{1}{l|}{EM-ICP   \cite{zhu2017event}}           & \multirow{3}{*}{Event}   & 0.161     & 0.120          & 0.337    & 0.334         \\
        \multicolumn{1}{l|}{HASTE   \cite{alzugaray2020haste}}      &    & 0.096     & 0.063          & 0.442    & 0.427         \\    
        \multicolumn{1}{l|}{AEB-Tracker \cite{wang2024asynchronous}}&    & 0.473     & 0.466          & 0.553    & 0.473         \\ 
        \multicolumn{1}{l|}{ETAP   \cite{hamann2024event}}      &    & 0.701     & 0.601          & 0.881    & \textbf{0.876}         \\  \hline
        \rowcolor{mygray}
        \multicolumn{1}{l|}{Ours}     & Event & \textbf{0.713}     & \textbf{0.626}          & \textbf{0.885}    & 0.875         \\ 
        \hline
        \Xhline{2.\arrayrulewidth}
        \end{tabular}
        \label{tab:ec_eds}
    }
\end{table}

\cref{sec4:imple} details the experimental setup. The next two sections comprehensively evaluate the proposed method on two tasks: tracking of any point and feature tracking. The key difference is that feature tracking focuses on high-contrast points using local descriptors, while tracking of any point extends this ability to low-texture regions. Due to the lack of prior work on event-based tracking of any point, comparisons are made with state-of-the-art video-based methods and their direct adaptation to the event domain. Feature tracking is compared with existing event-based methods and video-based TAP method adapted to this task, as feature tracking can essentially be seen as a specialized subset of the more general tracking any point problem.

\subsection{Implementation Details}
\label{sec4:imple}
Event time surface is generated by normalizing event timestamps and encoding them into two separate channels for positive and negative polarity events. The temporal window length is set according to the ground truth frequency, and timestamps within each window remain untruncated to fully preserve the motion trajectory over time. 

The model is trained on the Ev-PointOdyssey dataset at a spatial resolution of $512 \times 640$. Optimization is performed using the AdamW optimizer \cite{loshchilov2017decoupled} and OneCycleLR scheduler \cite{smith2019super} with a maximum learning rate of $5e-4$ and a cycle percentage of $0.1$. Weighted Mean Absolute Error loss is used for supervision, defined as: 
\begin{align}
    L = \sum_{i} \gamma^{T-i} \cdot \lvert y_{pred}^i - y_{true}^i \lvert, i = 1, \dots,T ,
\label{eq:loss}
\end{align}
where $\gamma$ is the weight factor, set to 0.8, and $T$ represents the number of bins per input sequence, set to 48. The purpose of this weighting is to impose a greater penalty on errors in later time steps, thereby improving the performance on long-term tracking.
Experiments are conducted in parallel on 8 Nvidia RTX A6000 GPUs, implemented in PyTorch.

\begin{figure*}[!t]
    \centering
    \includegraphics[width=\linewidth]{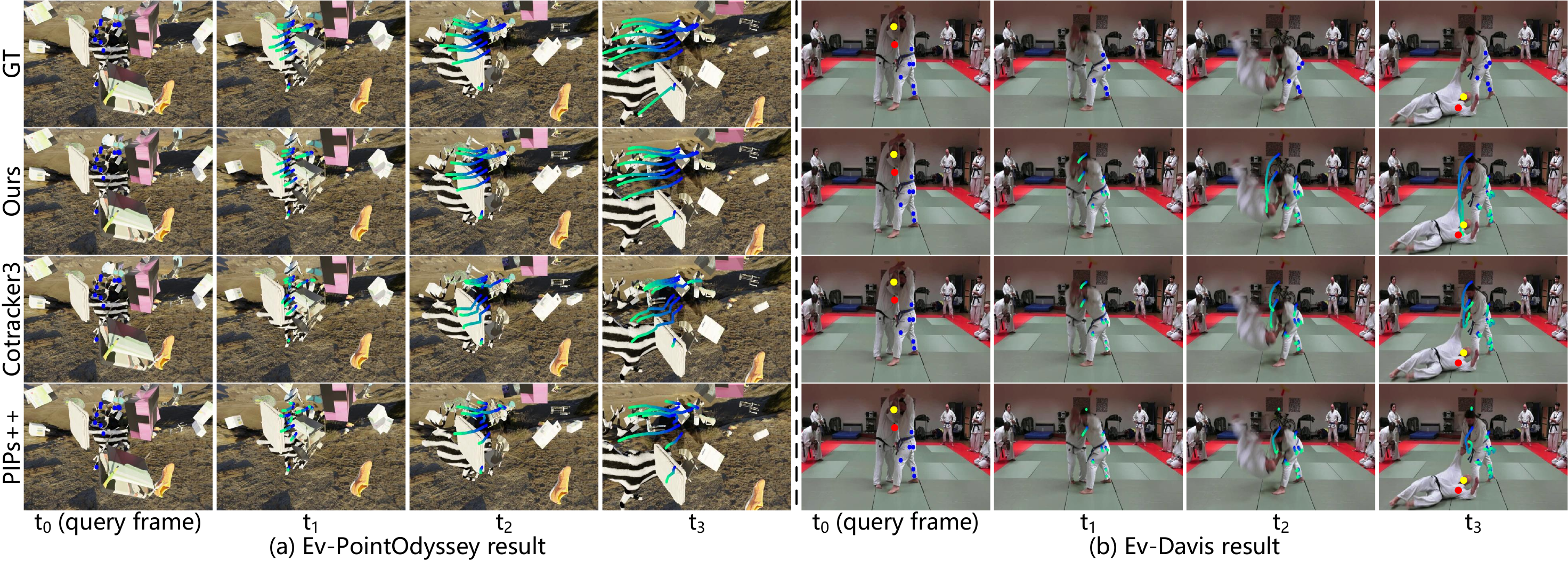}
    \caption{Qualitative results for Ev-PointOdyssey and Ev-Davis datasets. (a) The predicted trajectories are visualized using a blue-to-green colormap. By leveraging kinematic vectors for matching instead of relying solely on appearance features, the proposed method exhibits strong robustness to occlusion. (b) Due to missing ground truth for occluded points in the Ev-Davis dataset, only their positions are visualized. Notably, the proposed method leverages the high temporal resolution of event cameras to maintain stable tracking during fast motion (see red and yellow points), while video-based methods suffer from motion blur, leading to performance degradation.}
    \label{fig:methodVS}
\end{figure*}

\subsection{Task1: TAP}

\textbf{Dataset}.
To validate the effectiveness of the proposed method, this paper simulates events from the PointOdyssey dataset \cite{zheng2023pointodyssey} and TAP-Vid-Davis dataset \cite{doersch2022tap}, referred to as Ev-PointOdyssey and Ev-Davis. Compared to previous datasets \cite{doersch2022tap, harley2022particle}, PointOdyssey offers longer durations and more annotated points on average, with each video lasting approximately 2000 frames and 68 seconds on average, whereas prior datasets contain around 60 frames per video and last about 2 seconds. TAP-Vid-Davis is a real-world captured dataset, featuring faster motion and frequent point occlusions in the videos.
% In practice, event generation requires a continuous visual signal, which is typically achieved by rendering high-frame-rate videos for seamless flow. Here, the method from \cite{niklaus2020softmax} is applied to minimize pixel displacement between frames. Subsequently, DVS-voltemeter \cite{lin2022dvs} is utilized to synthesize realistic event data. Some examples from the two datasets are shown in \cref{fig:Ev-PointOdyssey}.

\textbf{Metrics}.
This paper follows the setup of \cite{zheng2023pointodyssey}, using $\delta_{avg}$, MTE, and $Survival_{50}$ for evaluation. $\delta_{avg}$ measures the percentage of trajectories within error thresholds of $\{1, 2, 4, 8, 16\}$, averaged across these values. Median Trajectory Error (MTE) calculates the distance between predicted and ground truth, using the median to reduce the impact of outliers. $Survival_{50}$ indicates the average duration until tracking failure, expressed as a percentage of the sequence length. Failure is defined as an L2 distance exceeding $50$ pixels. Additionally, model parameters is reported for comparing the resource demands of different methods.

\textbf{Quantitative results}.
\Cref{tab:ev_pointodyssey} compares the proposed method with video-based approaches for TAP, such as PIPs++ \cite{zheng2023pointodyssey}, Cotracker \cite{karaev23cotracker} and Cotracker3 \cite{karaev2024cotracker3}. Additionally, due to the absence of publicly available code for event-based methods, some video-based methods are directly applied to the event domain for a fairer comparison. However, these video-based methods rely solely on local iterative appearance matching, making them prone to errors in large displacements or nonlinear motion. This issue is further amplified when applied to event data, as events lack texture information. In contrast, the proposed method integrates kinematic vectors to guide matching, improving robustness in these challenging scenarios. $Survival_{50}$ metric shows a 17.9\% improvement over methods that solely rely on event data. In addition, it exhibits highly competitive performance even when compared to approaches that utilize video input.
Moreover, the proposed method is parameter-efficient. This efficiency stems from two factors: first, event cameras focus solely on dynamic changes, thereby reducing visual redundancy and enabling content learning without complex networks; second, a transformer replaces MLPs \cite{zheng2023pointodyssey, weikang2023context} to effectively capture temporal dependencies. 

\textbf{Qualitative analysis}.
% To do
\Cref{fig:methodVS} illustrates a comparison of the proposed method with prior works on the Ev-PointOdyssey and Ev-Davis datasets. PIPs++ \cite{zheng2023pointodyssey} and Cotracker3 \cite{karaev2024cotracker3} take video input, while the proposed method takes event input. Trajectories are shown in a blue-to-green colormap, indicating point movement. Due to the lack of ground truth for occluded points in the Ev-Davis dataset, accurately visualizing point trajectories is difficult, so only the point positions are visualized.

The left panel (a) shows results on Ev-PointOdyssey, where the query point is initially visible, gets occluded as motion starts, then reappears. Video-based methods like PIPs++ and Cotracker3, relying solely on appearance matching, struggle under these conditions. In contrast, the proposed method mitigates matching ambiguities caused by occlusion by using kinematic vectors as guidance. The right panel (b) shows results on the Ev-Davis, where motion blur degrades video-based methods. Leveraging the high temporal resolution of event cameras, the proposed method demonstrates robust tracking in this challenging scenario.

\subsection{Task2: Feature Tracking}
\textbf{Dataset}.
The Event Camera (EC) dataset \cite{mueggler2017event} and Event-aided Direct Sparse Odometry (EDS) dataset \cite{hidalgo2022event} are commonly used to evaluate event-based feature tracking methods.The EC dataset includes $240 \times 180$ resolution event streams and videos recorded with a DAVIS240C sensor. The EDS dataset contains videos and events captured simultaneously using a beam splitter with spatial resolution $640 \times 480$. Notably, both datasets provide asynchronous ground truth for feature points at varying timestamps. To ensure fair comparisons, spline interpolation is applied to align the ground truth to a common timestamp. 
\begin{figure}[t]
    \centering
    \begin{subfigure}{0.24\linewidth}
        \includegraphics[width=\linewidth]{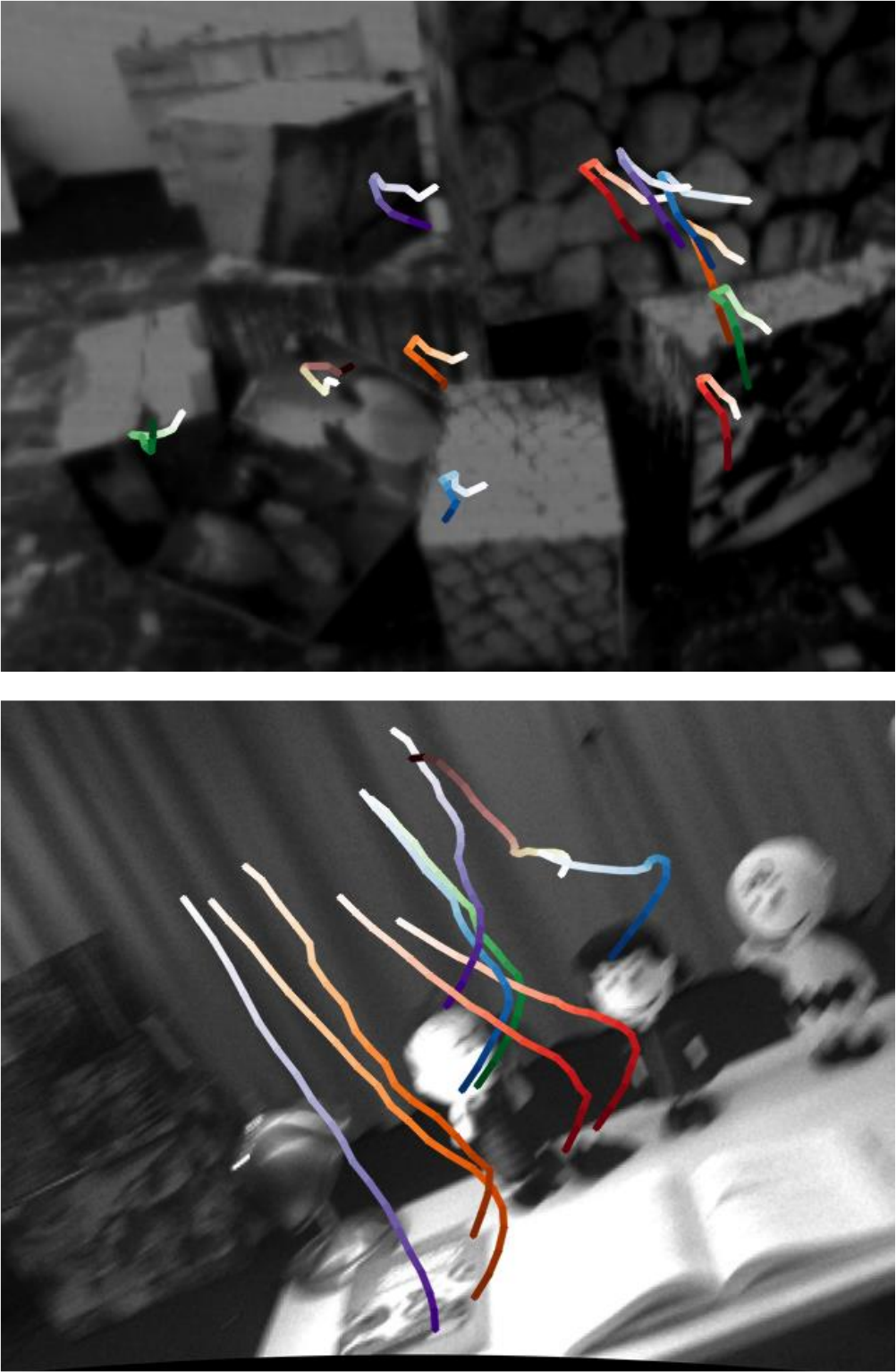}
        \caption{DeepEvT \cite{messikommer2025data}}
    \end{subfigure}
    \begin{subfigure}{0.24\linewidth}
        \includegraphics[width=\linewidth]{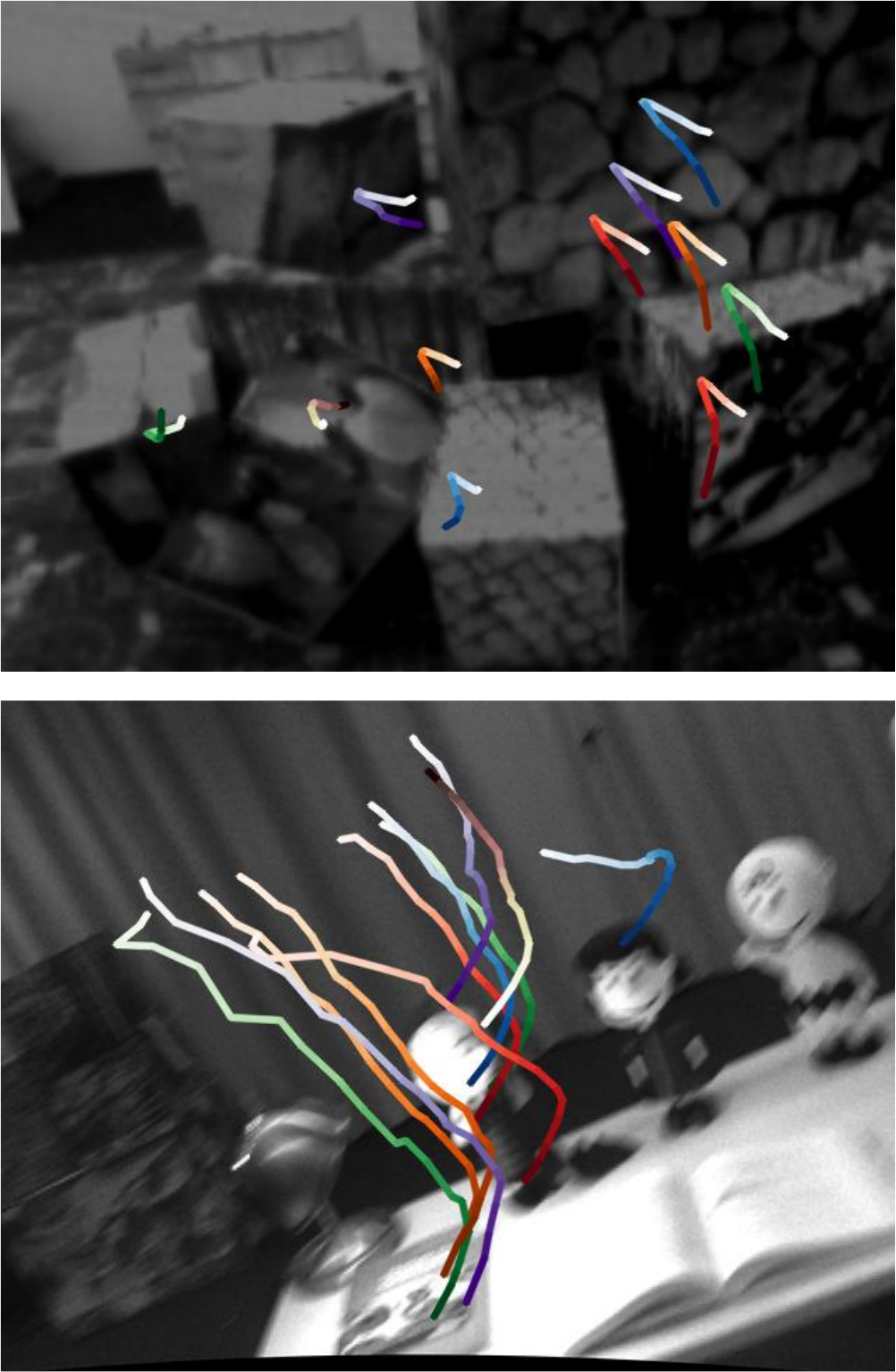}
        \caption{PIPs++ \cite{zheng2023pointodyssey}}
    \end{subfigure}
    \begin{subfigure}{0.24\linewidth}
        \includegraphics[width=\linewidth]{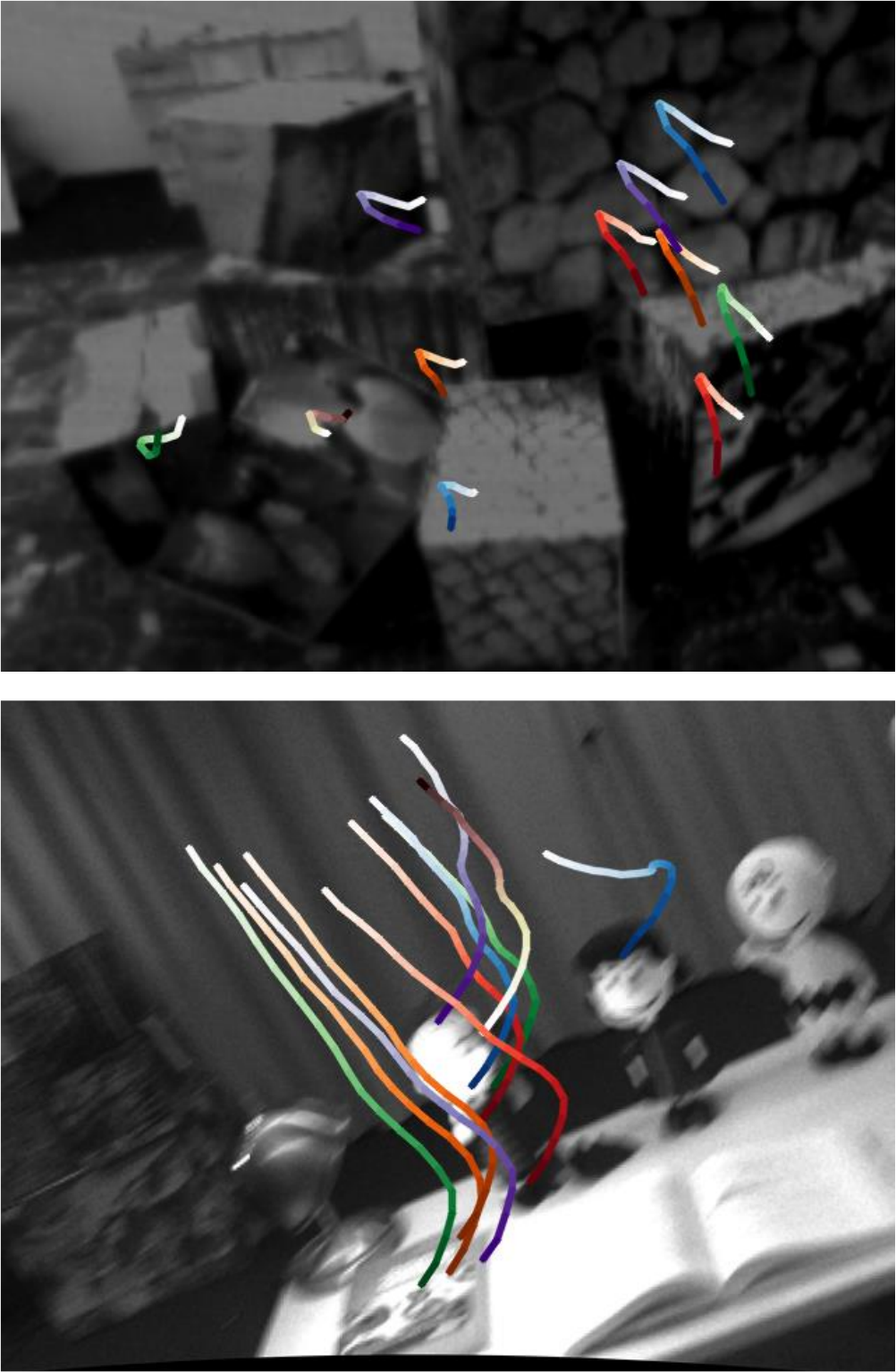}
        \caption{Ours}
    \end{subfigure}
    \begin{subfigure}{0.24\linewidth}
        \includegraphics[width=\linewidth]{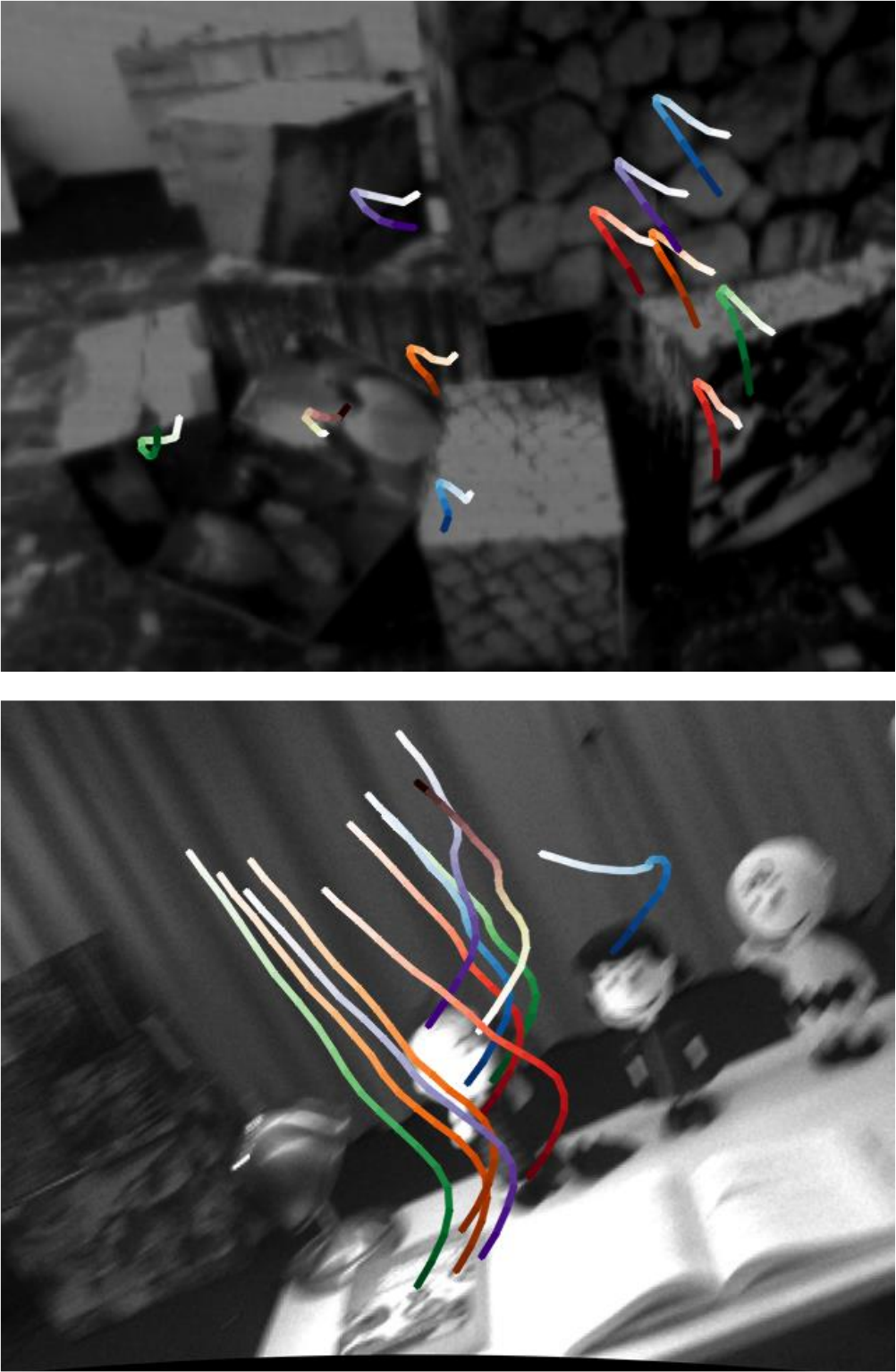}
        \caption{GT}
    \end{subfigure}
    \caption{Tracking results on EC (top row) and EDS (bottom row) dataset. DeepEvT fails to retain points under occlusion, while PIPs++ experiences performance degradation during rapid motion.}
    \label{fig:ec_result}
\end{figure}
 
\textbf{Metric}.
Quantitative metrics follow the setup of \cite{messikommer2023data}, utilizing Feature Age (FA) and Expected Feature Age (EFA). FA measures the percentage of successful tracking steps across thresholds from $1$ to $31$, with the final score being the average across all thresholds. EFA quantifies the lost tracks by calculating the ratio of stable tracks to ground truth and scaling it by the feature age.

\textbf{Results}.
The proposed method is compared against three categories of approaches across two datasets. First, video-based methods for tracking of any point, such as PIPs++ \cite{zheng2023pointodyssey} and Cotracker3 \cite{karaev2024cotracker3}. On the EDS dataset, motion blur caused by rapid movement degrades their performance, while on the EC dataset, low resolution leads to texture blur, further affecting their accuracy.
Second, methods that utilize both video frames and event data, including EKLT \cite{gehrig2020eklt}, DeepEvT \cite{messikommer2025data}, and FE-TAP \cite{liu2024tracking}. Among these, FE-TAP integrates events and video for tracking of any point. Since its implementation is not publicly available, the results reported in the original paper are used for a fair comparison.
Lastly, event-based tracking methods, such as EM-ICP \cite{zhu2017event}, HASTE \cite{alzugaray2020haste}, AEB-Tracker \cite{wang2024asynchronous} and ETAP \cite{hamann2024event}, are also evaluated. Across both datasets, the proposed method outperforms all three categories of existing trackers, see \cref{tab:ec_eds}.  \Cref{fig:ec_result} provides a visual comparison, showing that our method better maintains consistency under challenging conditions by leveraging motion cues and temporal modeling.

\begin{table}[t]
\centering
\caption{Ablation studies on each part of the proposed method.}
\resizebox{\linewidth}{!}{
    \begin{tabular}{ccccc|ccc}
    \hline
    \Xhline{2.\arrayrulewidth}
    \multicolumn{2}{c}{MGM} & \multicolumn{3}{c|}{VMA}                   & \multicolumn{3}{c}{Ev-PointOdyssey}    \\ \hline
        PF    & MLP                     & CA & TC & TA          & $\delta_{avg}\uparrow$  & MTE $\downarrow$       & $Survival_{50} \uparrow$     \\ 
    \hline
    \Xhline{2.\arrayrulewidth}
    \multicolumn{1}{c}{}   &       &        &           &                              & 0.312 & 43.64   & 0.488       \\
    \multicolumn{1}{c}{\checkmark} &       &        &   &                              & 0.325 & 38.72   & 0.485        \\
    \multicolumn{1}{c}{\checkmark} & \checkmark  &    &   &                            & 0.340 & 35.65   & 0.513        \\ \hline
    \multicolumn{1}{c}{\checkmark} & \checkmark     & \checkmark & &                   & 0.360 & 29.56   & 0.538        \\
    \multicolumn{1}{c}{\checkmark} & \checkmark & \checkmark & \checkmark &            & 0.358 & 27.63   & 0.549        \\
    \rowcolor{mygray}
    \multicolumn{1}{c}{\checkmark} & \checkmark & \checkmark & \checkmark & \checkmark & \textbf{0.362} & \textbf{25.63}   & \textbf{0.560}   \\     
    \hline
    \Xhline{2.\arrayrulewidth}
    \end{tabular}
}
\label{tab:ablation}
\end{table}
\begin{figure}[t]
    \centering
    \begin{subfigure}{0.32\linewidth}
        \includegraphics[width=\linewidth]{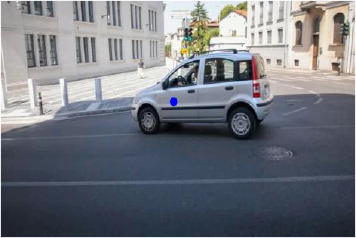}
        \caption{Query frame}
    \end{subfigure}
    \begin{subfigure}{0.32\linewidth}
        \includegraphics[width=\linewidth]{figs/ablationMGM_3.pdf}
        \caption{w/o MGM}
    \end{subfigure}
    \begin{subfigure}{0.32\linewidth}
        \includegraphics[width=\linewidth]{figs/ablationMGM_4.pdf}
        \caption{w/ MGM}
    \end{subfigure}
    \caption{Effectiveness of MGM. (a) The initial frame containing the query point. (b) Results from appearance-based matching without MGM. (c) Results incorporating kinematic vectors extracted by MGM for enhanced matching.}
    \label{fig:ablation_mgm}
\end{figure}

\subsection{Ablation Study}
\textbf{Impact of the motion-guidance module}. The first three rows of \cref{tab:ablation} illustrate how kinematic vectors from MGM contribute to point displacement updates, where PF stands for Plane Fitting. The first row shows a baseline without MGM, relying solely on appearance matching. The second row includes MGM but omits MLP-based correction for ambiguous kinematic vectors. Results indicate that kinematic vectors play a significant role in enhancing tracking accuracy, and this effect is further amplified when corrected by the MLP. As shown in \cref{fig:ablation_mgm}, MGM is essential for handling low-texture regions, where appearance-based matching alone often fails due to the lack of texture information in event data. By introducing kinematic vectors as additional motion cues, MGM effectively compensates for this limitation, improving robustness in challenging scenarios.

\begin{figure}[t]
    \centering
    \begin{subfigure}{0.48\linewidth}
        \includegraphics[width=\linewidth]{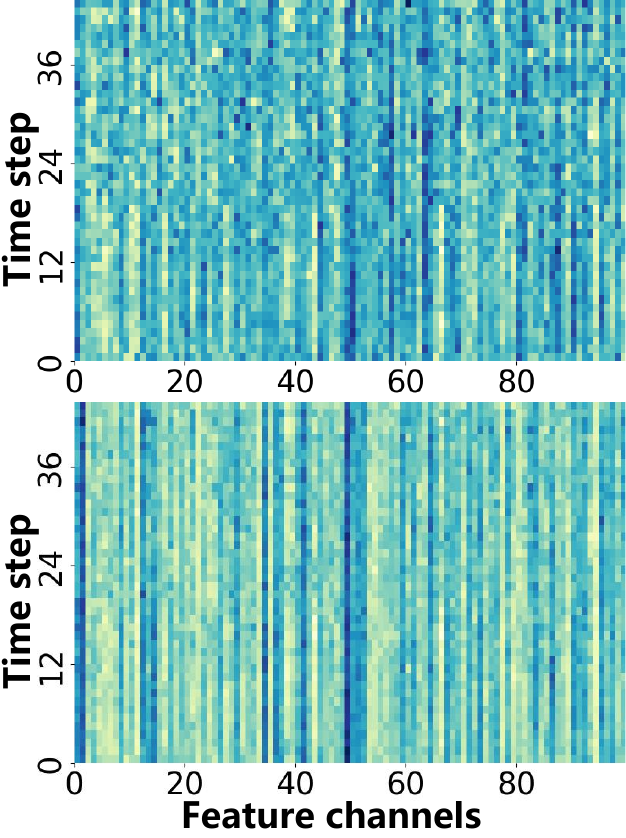}
        \caption{Temporal distribution}
    \end{subfigure}
    \begin{subfigure}{0.48\linewidth}
        \includegraphics[width=\linewidth]{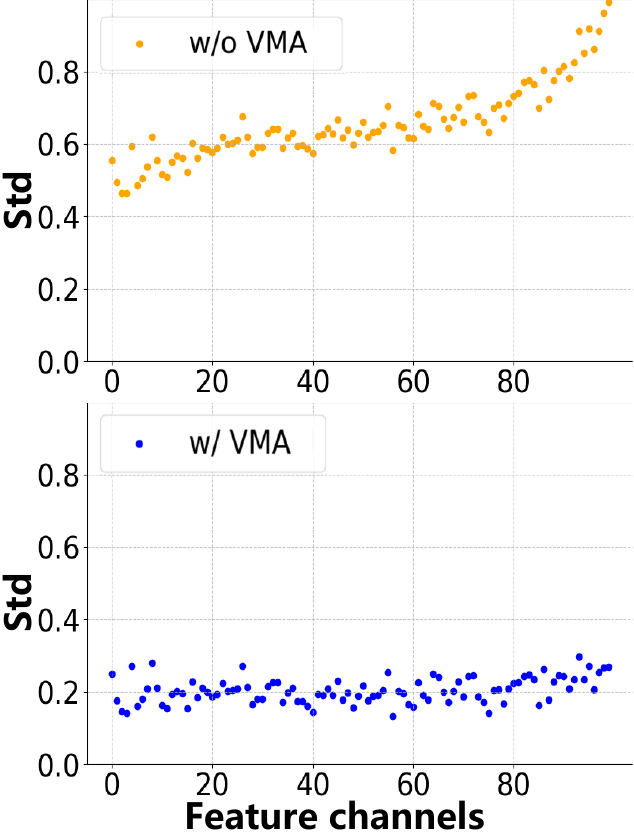}
        \caption{Standard deviation}
    \end{subfigure}
    \caption{Effectiveness of VMA. The first row shows results using only appearance features, without VMA, while the second row includes VMA. (a) Temporal distribution across feature channels. (b) Temporal standard deviation across feature channels.}
    \label{fig:ablation_vam}
\end{figure}

\textbf{Effectiveness of variable motion aware module}. Rows four through six in \cref{tab:ablation} present the ablation studies for each component of VMA, where CA, TC, and TA represent Cross Attention, Temporal Convolution, and Temporal Attention, respectively. The fourth row performs appearance matching guided by kinematic vectors. The fifth row employs 1D temporal convolutions to capture short-term dependencies. The final row retains the full setup, modeling temporal relationships via both short- and long-term paths. Results show that integrating kinematic guidance and temporal modeling progressively enhances performance, demonstrating the effectiveness of each VMA component. \Cref{fig:ablation_vam} offers a visual comparison. Features without CA, TC, and TA (\ie, $\mathbf{G}^k$) show temporal discontinuities, while enhanced features exhibit improved consistency and lower standard deviation.

\section{Conclusion}
\vspace{-0.5em}
\label{sec:conclusion}
This study introduces a novel event-based framework for tracking any point, leveraging a motion-guidance module to extract kinematic features that refine the matching process and constructs a dynamic-appearance space. Additionally, the integration of a variable motion aware module enables the system to account for motion variations, ensuring temporal consistency across diverse velocities. To evaluate the approach, two simulated event point tracking dataset was collected. The proposed method outperforms existing approaches on both tracking of any point and feature point tracking tasks across four datasets, with competitive model parameters. This technique holds substantial potential for applications in embodied intelligence and related fields.
% This technique holds substantial potential for applications in embodied intelligence and related fields.
% This technique holds substantial potential for applications in embodied intelligence, autonomous driving, and related fields.
\vspace{-0.5em}
\section*{Acknowledgment}
\vspace{-0.5em}
\noindent
\begin{minipage}[t]{1\linewidth}
This work is supported by the National Natural Science Foundation of China (NSFC) under Grants 62225207, 62436008 and 62306295.
\end{minipage}

\clearpage
\setcounter{page}{1}
\maketitlesupplementary

\section{Network Structure}
\textbf{Network Encoder}. The encoder is based on ResNet \cite{he2016deep} and consists of four feature extraction modules, each containing two stacked residual blocks. The first block in each module performs downsampling, while the second maintains the resolution. Features at different resolutions from each module are aligned to a consistent resolution via interpolation and concatenated along the channel to form multi-scale representations. The concatenated features are then compressed along the channel using a $1 \times 1$ convolution.

% \textbf{Position embedding}. The point displacements between adjacent time steps provide apparent motion vectors that assist in iterative position updates. To leverage this, 2D sine-cosine encoding is applied to the forward and backward displacements, generating 128-dimensional feature representations. These motion features, concatenated with other features, serve as inputs to the transformer. The detailed network architecture is shown in Fig. 3 of the main paper.

\section{Implementation Details}
\textbf{Training details}.
To accelerate training, events in Ev-PointOdyssey are converted into time surface \cite{mueggler2017fast} representations at the original spatial resolution and stored. Sampling is conducted with step sizes of $\{1, 2, 4\}$, where each sampled sequence has a length of 48. Starting points for sampling are selected along the timeline at intervals of 2, ensuring comprehensive coverage of the entire sequence. This process generates approximately $340k$ training samples, each containing $128$ tracking points.

\textbf{Data augmentations}.
To improve model robustness, several data augmentation techniques are applied. Random pixel erasure is performed at all time steps, with varying erased regions to enhance the network's ability to handle occlusions. The erased values are replaced by the mean of the original region. Additionally, random scale-up is employed to improve tracking across different object scales, with scale parameters for the next time step initialized from the previous step and adjusted with slight random perturbations. Spatial flipping and temporal reversal are also applied to further diversify the dataset.
  
\section{Additional Ablation Experiments}
In addition to the ablation studies reported in the main paper, experiments are conducted on the input event representation and the number of pixels used for plane fitting in the motion-guidance module. Due to time constraints, these experiments are performed on a baseline model, which excludes the variable motion aware module, and are conducted at a lower resolution of $192 \times 256$.

\begin{table}[t]
\centering
\caption{Ablation studies on different input event representations.}
\resizebox{\linewidth}{!}{
    \begin{tabular}{c|ccc}
    \hline
    \Xhline{2.\arrayrulewidth}
    \multirow{2}{*}{Representations} & \multicolumn{3}{c}{Ev-PointOdyssey}    \\ \cline{2-4}
                    & $\sigma_{avg}\uparrow$  & MTE $\downarrow$       & $Survival_{50} \uparrow$     \\ 
    \hline
    \Xhline{2.\arrayrulewidth}
    Event image   & 0.298  & 54.43    & 0.443        \\
    Voxel grid    & 0.312  & 51.24    & 0.457        \\
    Time surface  & 0.323  & 48.41    & 0.477        \\ \hline
    \Xhline{2.\arrayrulewidth}
    \end{tabular}
}
\label{tab:suppl_ablation_input}
\end{table}
\begin{figure}[t]
    \centering
    \begin{subfigure}{\linewidth}
        \includegraphics[width=\linewidth]{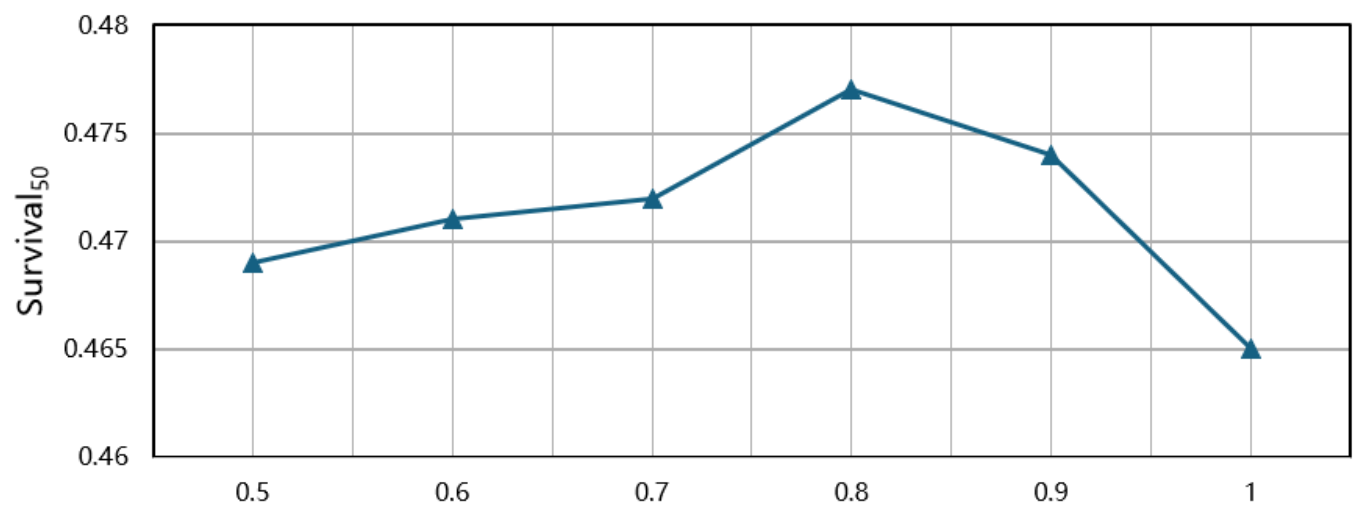}
        \caption{Ablation study on $\gamma$}
        \label{gamma}
    \end{subfigure}
    \begin{subfigure}{\linewidth}
        \includegraphics[width=\linewidth]{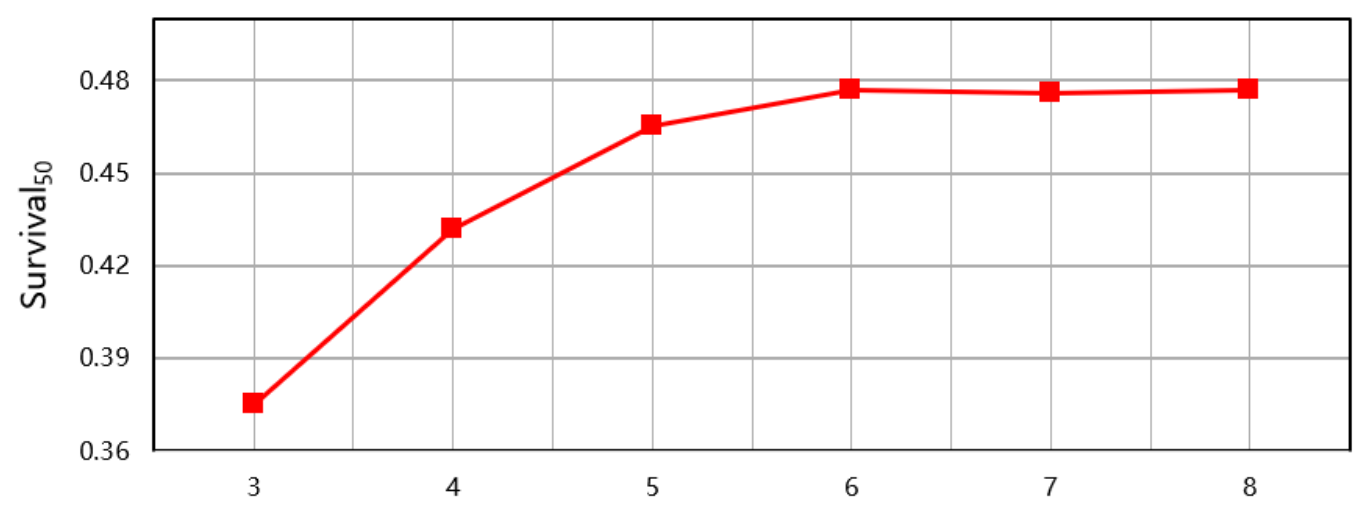}
        \caption{Ablation study on iterations}
        \label{iter}
    \end{subfigure}
    \caption{The survival metric is maximized when gamma and iteration are set to 0.8 and 6, respectively.}
    \label{fig:ablation_itera}
\end{figure}

\textbf{Input representation}. 
Different input event representations significantly impact event-based networks. To evaluate their effects on this task, various representations are used to train the baseline model, with the results presented in \cref{tab:suppl_ablation_input}. Event image  aggregates the number of events at each pixel over a fixed time window. Voxel grid \cite{zhu2019unsupervised} represents a spatiotemporal histogram, where each voxel corresponds to a specific pixel and time interval. Time surface \cite{mueggler2017fast} is a 2D map where each pixel stores the timestamp of the most recent event. As highlighted in the introduction of main paper, the TAP task relies on continuous motion information. Both event image and voxel grid quantize event timestamps, discarding dense temporal details, which limits their ability to capture smooth motion dynamics. In contrast, time surface retains rich temporal information, with each pixel encoding the motion history at its location, making it more suitable for encoding temporal motion cues effectively. As a result, time surface demonstrates superior performance compared with other representations.
\begin{figure*}[t]
    \centering
    \includegraphics[width=\linewidth]{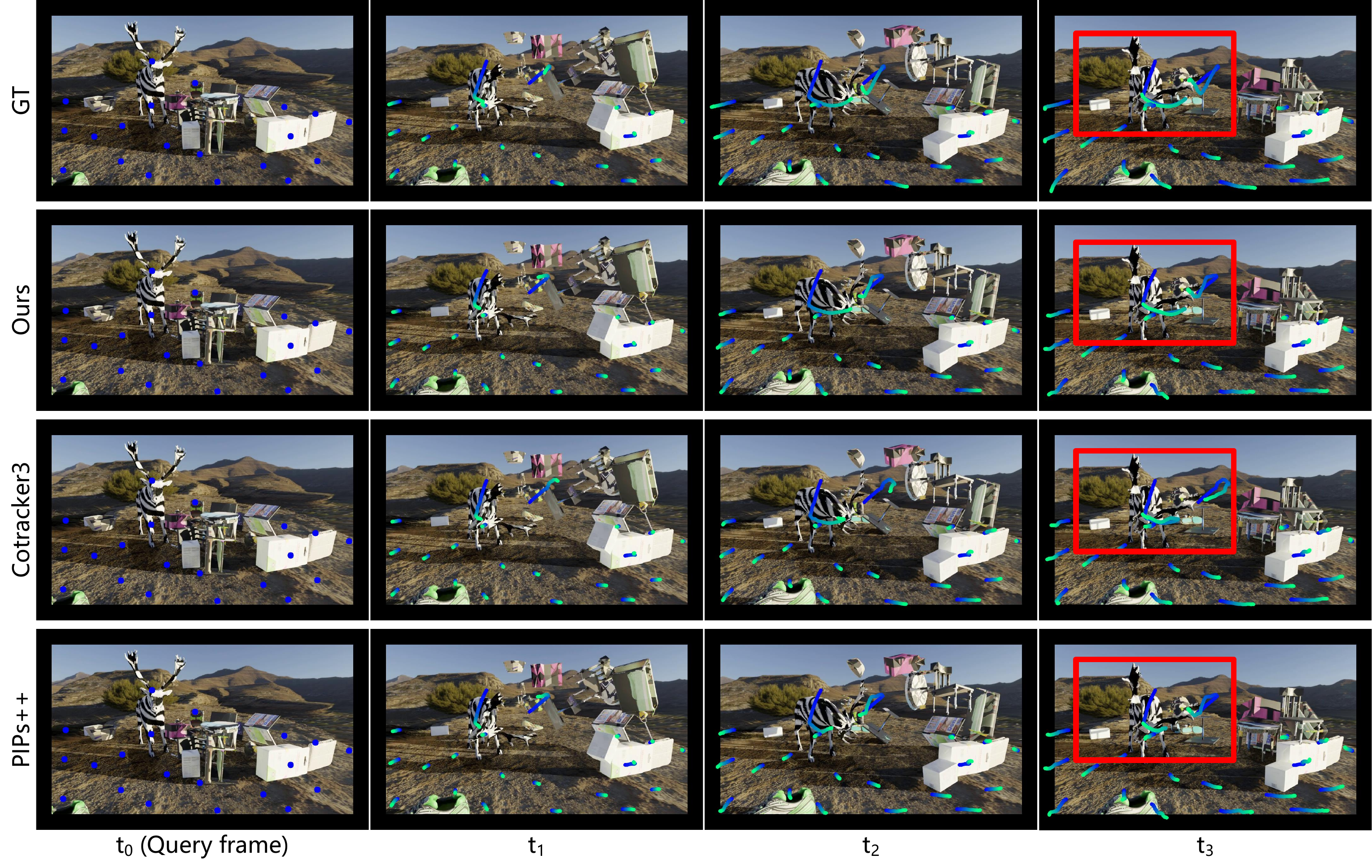}
    \caption{Qualitative results for Ev-PointOdyssey dataset.}
    \label{fig:add_qua}
\end{figure*}

\begin{figure}[t]
    \centering
    \includegraphics[width=\linewidth]{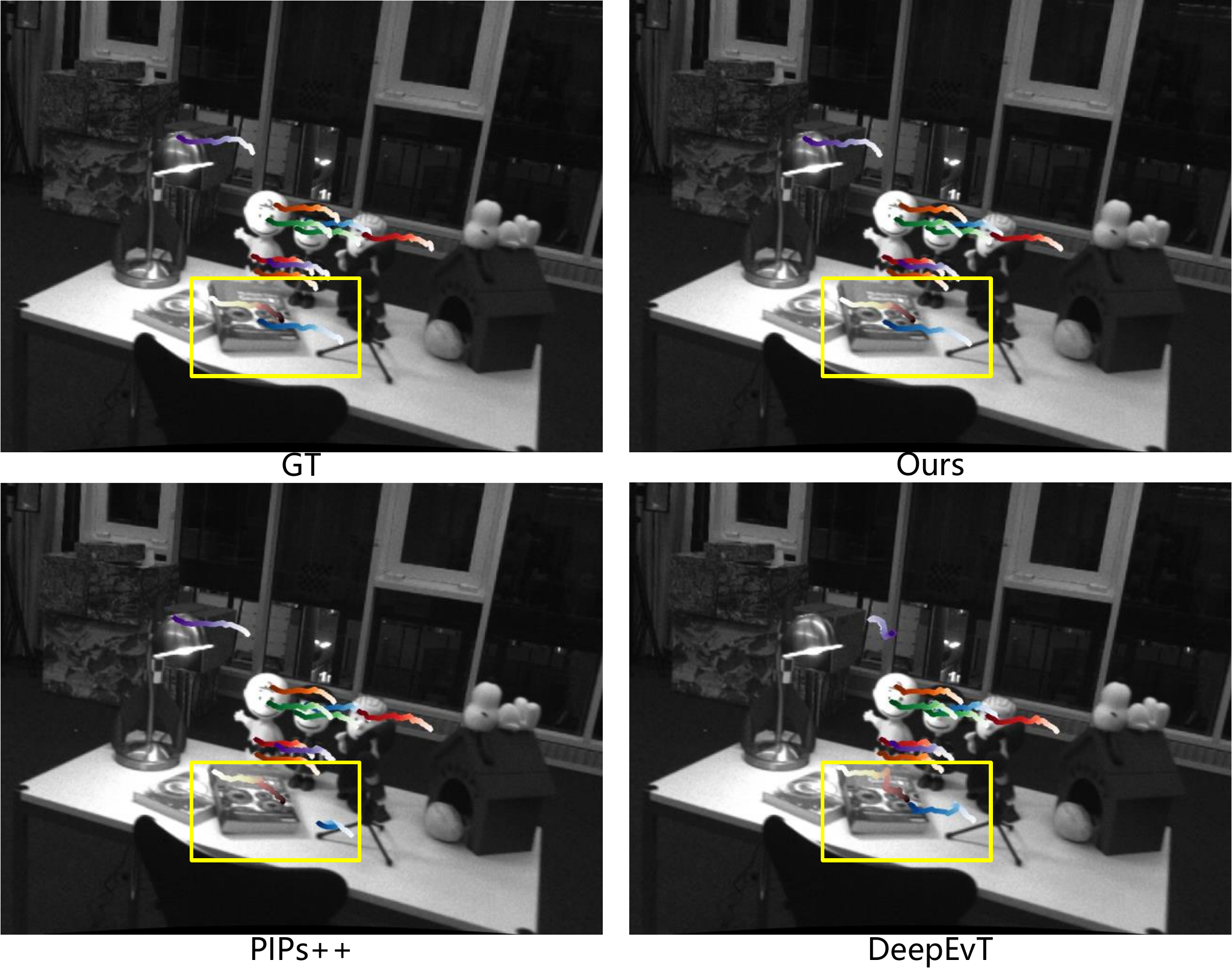}
    \caption{Qualitative results for EDS dataset.}
    \label{fig:add_eds}
\end{figure}

\textbf{Weight factor $\gamma$ in loss function}.
As described in Eq. (8) of the main paper, the losses across different time steps are weighted by a factor $\gamma$, assigning a higher penalty to losses occurring at later time steps. An ablation study is performed to identify the optimal $\gamma$. \Cref{gamma} demonstrates that the $Survival_{50}$ is maximized when $\gamma$ equals 0.8.

\textbf{The number of iterations}.
In addition to $\gamma$, the number of iterations also significantly affects tracking accuracy. While increasing the iteration count typically enhances accuracy, it also requires more computational resources. To find an optimal trade-off, an ablation study on the iteration count is performed. As illustrated in \cref{iter}, when the iteration count reaches 6, further increases do not yield significant improvements in the metrics.

% \textbf{Pixel count for plane fitting in MGM}.
% Estimating the kinematic features of a point requires constructing a local gradient plane using its neighboring pixels, where the number of sampled points significantly impacts the fitting quality. To determine the optimal number of points, an ablation study is conducted. Since at least four points are required for plane fitting, the experiment starts with four uniformly sampled points within a radius of $1$ pixel around the target. As shown in \cref{tab:suppl_ablation_pixel}, using four points achieves the best performance. This is because: (1) in smooth gradient regions without motion overlap, four points are sufficient to capture the gradient characteristics accurately; (2) in regions with motion overlap, additional points may introduce conflicting gradients, leading to deviations in the plane parameters and reduced fitting accuracy. 

\section{More Qualitative Results}
Additional qualitative results are provided in \cref{fig:add_qua,fig:add_eds} with higher-resolution images for clearer visualization. 
% Additional qualitative results are presented in \cref{fig:add_qua}. Our method establishes accurate correspondences in fast motion and low-texture regions, while demonstrating robustness to occlusions and boundary crossings.
{
    \small
    \bibliographystyle{others/ieeenat_fullname}
    \bibliography{reference}
}


\end{document}